\documentclass{article}
\usepackage[preprint]{spconf,amsmath,graphicx}
\usepackage{MnSymbol}
\usepackage{wasysym}
\usepackage{hyperref,url,latexsym,color,soul}

\copyrightnotice{\copyright IEEE 2021}
\toappear{To appear in {\it Proc. ICASSP2021, June 06-11, 2021, Toronto, Canada}}

\newcommand{\hc}[1]{#1}
\newcommand{\mf}[1]{#1}
\newcommand{\sml}[1]{#1}

\title{Machine Translation Verbosity Control for Automatic Dubbing}

\name{\normalsize  Surafel M. Lakew, Marcello Federico, Yue Wang,$^{\dagger}$\thanks{$^{\dagger}$Author carried out the work during an internship at Amazon.} 
Cuong Hoang, Yogesh Virkar, Roberto Barra-Chicote, Robert Enyedi}

\address{{\normalsize Amazon AI} \\
{\small \tt \{surafelm|marcfede|hoacuong|yvvirkar|rchicote|renyedi\}@amazon.com}
}

\begin{document}
\ninept
\maketitle
\begin{abstract}
Automatic dubbing aims at seamlessly replacing the speech in a video document with synthetic speech in a different language. The task implies many challenges, one of which is generating translations that not only convey the original content, but also  match the  duration of the corresponding utterances. In this paper, we focus on the problem of controlling the verbosity of machine translation output, so that subsequent steps of our automatic dubbing pipeline can generate dubs of better quality. We propose new methods to control the verbosity of MT output and compare them against the state of the art with both intrinsic and extrinsic evaluations. For our experiments we use a public data set to dub English speeches into French, Italian, German and Spanish. Finally, we report extensive subjective tests that measure the impact of MT verbosity control on the final quality of dubbed video clips. 
\end{abstract}
\section{Introduction}
\label{sec:intro}
Automatic Dubbing (AD) is the task of automatically replacing the speech in a video document with speech in a different language, while preserving as much as possible the user experience of the original video. AD differs from  speech translation \cite{casacuberta_recent_2008,sperber_speech_2020} in significant ways. In speech translation, a speech utterance in the source language is recognized, translated (and possibly synthesized) in the target language.
In speech translation, close to real-time response is expected and typical use cases include human-to-human interaction, traveling, live lectures, etc. Corresponding human tasks, from which data can be catered,  are  consecutive and simultaneous interpretation,  where either isolated sentences or a continuous stream of speech are translated. 
On the other hand, AD tries to automate the localization of audiovisual content, a complex and demanding workflow \cite{chaume:2004} managed during post-production by dubbing studios.

A major requirement of dubbing is speech synchronization which, in order of priority, should happen at the utterance level (isochrony), lip movement level (lip synchrony) and body movement level (kinesic synchrony) \cite{chaume:2004}. Most of the work on AD \cite{oktem2019,federico_speech--speech_2020,federico_evaluating_2020}, including this one,  addresses isochrony, aiming to generate translations and utterances that match the phrase-pause arrangement of the original audio. Hence, given the transcript of a source speech utterance, provided with time stamps, the first step is to generate a translation that fits the duration of the original utterance \cite{saboo_integration_2019,Lakew19}. Then, a {\em prosodic alignment}  \cite{oktem2019,federico_speech--speech_2020,federico_evaluating_2020} step follows, segmenting the translation into phrases and pauses corresponding to the phrases and pauses present \sml{in} the original speech. Finally, the sequence of phrases and pauses is passed to a text-to-speech synthesizer that generates each phrase by adjusting the speaking rate to fit its required duration.

This paper focuses on the MT step, controlling the output length so that AD can produce utterances of the same duration of the original speech. 
In this work, we use the number of characters in the sentence as a proxy of the duration of its spoken realization.\footnote{We empirically found that characters work better for this purpose than syllables computed with the tool used in \cite{saboo_integration_2019}.} 
Hence, we focus on controlling the number of characters in the MT output to be more or less the same as that of the source sentence.  Our work is inspired by recent works that have addressed verbosity control for text summarization \cite{kikuchi_controlling_2016,fan_controllable_2017,takase_positional_2019}, of neural MT models \cite{niehues_machine_2020, agrawal_controlling_2019, marchisio_controlling_2019, saboo_integration_2019,Lakew19} and ways to condition MT to side information \cite{hoang_improved_2018,michel-neubig-2018-extreme}. In particular, we implement and extend approaches proposed in \cite{saboo_integration_2019,Lakew19,hoang_improved_2018,michel-neubig-2018-extreme} to control the verbosity of MT for speech dubbing. As a significant difference to previous work we perform both intrinsic and extrinsic  evaluations of the generated translations, for which we use a publicly available data set \cite{MUSTC:2019} with speeches from English to French, Italian, German and Spanish.  Intrinsic evaluations measure MT quality and verbosity with respect to human post-edited translations matching length requirements, while extrinsic evaluations measure subjective quality of video clips dubbed by using the generated translations. To our knowledge, our work is the first to provide a systematic comparison of verbosity control methods in MT and a subjective evaluation that measures their usefulness on automatically dubbed content.

Our paper is arranged as follows. First, we describe the AD architecture used for our experiments. We then focus on existing and new methods for controlling the verbosity of MT output. Finally, we present and discuss experimental results of all compared methods.

\section{Dubbing Architecture}
\label{sec:AD}
\begin{figure}[t]
    \centering
    \includegraphics[width=\columnwidth]{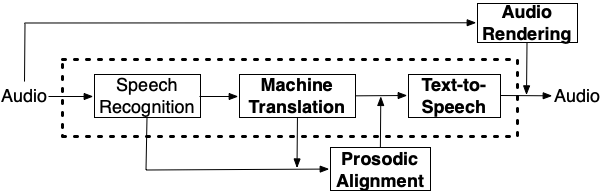}
    \caption{Speech translation pipeline (dotted box) with enhancements introduced  to perform automatic dubbing (in bold).}
    \label{fig:NTTS}
\end{figure}
This work builds on the AD architecture presented in \cite{federico_speech--speech_2020,federico_evaluating_2020} (Figure~\ref{fig:NTTS}) 
that extends a speech-to-speech translation \cite{casacuberta_recent_2008,weiss_sequence--sequence_2017,cross_vila_end--end_2018} pipeline with: neural machine translation (MT)  robust to ASR errors and able to control verbosity of the output \cite{vaswani2017attention,Lakew19,DiGangi19}; prosodic alignment (PA) \cite{oktem2019} which addresses phrase-level synchronization of the MT output by leveraging the force-aligned source transcript; neural text-to-speech (TTS) \cite{ntts01,ntts02,tts03} with precise duration control; and, finally, audio rendering that enriches TTS output with the original background noise (extracted via audio source separation with deep U-Nets \cite{ronneberger2015u,jansson2017singing}) and reverberation, estimated from the original audio  \cite{lollmann2010improved,habets2006room}.

\section{MT with Verbosity Control} 
\label{sec:TTS}
\mf{Automatic dubbing calls for translations which can be fluently uttered 
within the same time interval of the original source speech~\cite{saboo_integration_2019,Lakew19}. Given that text-to-speech can stretch its speaking rate without noticeable effects\footnote{This holds at normal speaking rates, but clearly not when the original speech has already extreme speaking rates.}, ideally we  would like MT to produce translations that are within a $\pm 10\%$ range of the original length, which we measure in number of characters. 
In the following, we present a range of approaches that we investigated to pursue this goal.}

\subsection{Naive Length Control}
\mf{The simplest way to control verbosity of MT is to end the inference once the output has reached the target length. However, with the natural difference in verbosity among languages, it is obvious that this sole criterion could lead to poor MT performance.} 
\mf{A better alternative is to leverage the already existing {\em length penalty} ~\cite{wu2016google} used at search time to avoid the NMT model producing too short or incomplete translations. It normalizes the log-prob scoring function as:
$$
S(t,s) = \frac{log P(t|s)}{LP(t)} + CP(s,t),
$$
where the coverage penalty ($CP$) penalizes translation that fully covers the source, whereas length penalty ($LP$) is ~\cite{wu2016google}:}
\begin{equation}
  LP(t) = (5 + |t|)^{\alpha}/(5+1)^{\alpha}.
  \label{eq:lenpenalty}
\end{equation}
\mf{Following~\cite{Lakew19}, we found that $\alpha=0.5$ provides the best trade-off between verbosity and translation quality.
}

\subsection{Verbosity Token}
\mf{As proposed by \cite{Lakew19}, we can introduce a special source token that specifies the desired verbosity in the translation. To train NMT to learn this behavior,  we first need to compute the target-source length ratio (LR) of all entries in the training data. Then, we categorize the training examples into three classes (\textit{Short}, \textit{Normal} and \textit{Long}) based on their LR as follows:}
\begin{equation}
  v =
    \begin{cases}
      {\tt Short} & \text{if $LR < 0.97$}\\
      {\tt Normal} & \text{if $0.97 \leq LR \leq 1.05$}\\
      {\tt Long} & \text{if $LR > 1.05$}.
    \end{cases}
    \label{eq:verbosity-tok}
\end{equation}
\begin{figure}[t]
    \centering
    \includegraphics[scale=0.95,trim={1.0cm 22.5cm 11.0cm 1.1cm},clip]{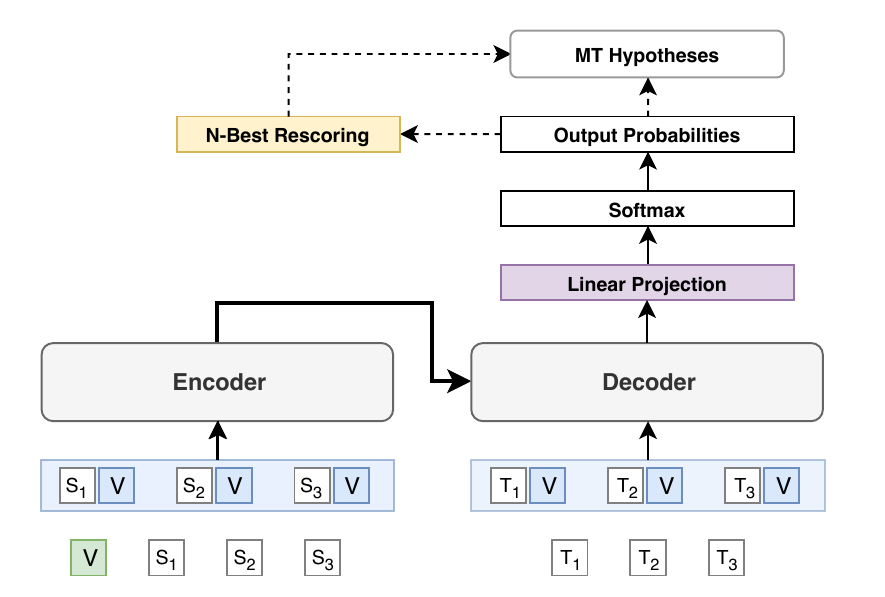}
    \caption{Our MT modeling variants with verbosity control: (green) verbosity token ($v$) added to input source, (blue) summing $v$ on the source/target embeddings, (violet) adding $v$ as an additional bias and (yellow) rescoring output hypotheses.}
    \label{fig:components}
\end{figure}
\mf{At training time, the verbosity token ($v$) is assigned to an embedding vector like any other token of the source vocabulary.}
\mf{Formally, we feed the MT encoder a sequence of embeddings as follows:}
\begin{align}
E_{source} = [E(v), E(tok_1), ..., E(tok_N)].
\label{eq:verb-tokseq}
\end{align}
\mf{Where $E(\dot)$ is the embedding lookup function and $N$ is the number of tokens in the source sentence.
The model is trained end-to-end so that both $E(v)$ and all MT parameters are jointly learned.} 

\noindent
\mf{At inference time, we prepend a desired $v$ value to the source sequence (e.g. {\tt Normal}). This encourages the MT model to generate translations that are within the corresponding LR range of Eq.~\ref{eq:verbosity-tok}.}

\mf{Although the verbosity token approach has shown to be effective \cite{Lakew19}, we explore further means to inject verbosity information in the model and in the search process (see Figure~\ref{fig:components}).}

\subsection{Verbosity Embeddings}
\mf{Verbosity information, once mapped to an embedding,  can  be integrated into the encoder and decoder in various ways.}

\noindent
\subsubsection*{Summing Verbosity to Token Embeddings}
As an alternative to (\ref{eq:verb-tokseq}), we can feed the encoder with the sequence:
\begin{align}
E_{source} = [E(tok_1)+E(v), ..., E(tok_N)+E(v)].
\label{eq:verb-emb-enc-dec}
\end{align}
Trivially, the same idea can be also applied to the input of the decoder. We will thus experiment with all three combinations: only encoder, only decoder, both encoder-decoder. Another alternative we consider, is to use the verbosity encoding and in addition the verbosity embedding in the decoder. \mf{Our motivation  is to reinforce the influence of the verbosity token in MT and investigate whether this scheme makes any additional impact on controlling MT output verbosity without sacrificing translation quality.}

\noindent
\subsubsection*{Verbosity as Output Layer Bias}
\mf{We can use the verbosity embedding $E(v)$ as an extra bias vector ~\cite{michel-neubig-2018-extreme} in the final linear projection layer of the decoder: 
\begin{equation}
O_t = W S_t + b + E(v),
  \label{eq:verb-emb-output}
\end{equation}
where $S_t$ is the decoder state at time $t$, $W$ and $b$ are the transformation and bias vector of the output layer and $O_t$ is the output vector.}

\subsection{Fine-Tuning with Verbosity Information}
\mf{It is suggested by \cite{Lakew19}  to apply training of the verbosity token as a fine-tuning stage of a pre-trained model. This work explores this direction further and compares two fine-tuning approaches}:
\begin{itemize}
\item Single-stage fine-tuning: Fine-tune a generic model trained with large scale generic data with in-domain data augmented with verbosity token, as in ~\cite{Lakew19}.
\item Two-stage fine-tuning: Fine-tune the generic model with verbosity information on generic data and then fine-tune again on in-domain data with verbosity information.
\end{itemize}

\subsection{Rescoring Translation Output}
\mf{In order to generate translations suited for dubbing, ~\cite{saboo_integration_2019} proposed to rescore N-best translations ($t$) generated with a large beam size ($B$) with the following function:}
\begin{equation}
  S_d(t,s) = (1-\alpha) \log P(t\mid s) + \alpha S_p(t,s),
    \label{eq:nbest-rescoring}
\end{equation}
\mf{where $S_p$ is the synchrony score computed by:\footnote{With the minor difference that in \cite{saboo_integration_2019} the length is expressed in syllables while here in characters.}}  
\begin{equation}
S_p (t,s) = (1+|len(t)-len(s)|)^{-1}.
    \label{eq:nbest-rescoring-diff}
\end{equation}
\mf{The factor $\alpha$ is adjusted to set the relative importance of length-similarity versus translation-probability.  As reported in~\cite{saboo_integration_2019} and confirmed by our experiments, for high $\alpha$ values the synchrony sub-score ($S_p$) can cause significant performance drop.

In this work, we hypothesize that it is suboptimal to use the synchrony sub-score as in Eq.~\ref{eq:nbest-rescoring-diff} because it aims to make long output shorter and short output longer at the same time. This is not necessary because in practice, we often find the need of either reducing or increasing the length ratio (LR). More specifically, translation directions in our experiments are from \textit{English} to other languages. To make MT output more or less the same as that of the source sentence, we often find the need of reducing the LR and not the other way around. To this end, we propose a new unidirectional version of the synchrony sub-score that encourages the decrease of the LR of the translations during rescoring:}
\begin{equation}
S_p (t,s) = (1+\frac{len(t)}{len(s)})^{-1}.
    \label{eq:nbest-rescoring-ratio}
\end{equation}
\mf{Compared to the original synchrony sub-score, our proposed scoring function is not only simpler but also fits better to the need of reducing the LR. As a side note, for the opposite translation directions (from other languages to \textit{English})  we can simply reverse the LR of the translations during rescoring.}

\begin{figure*}[t]
    \centering
    \includegraphics[width=2\columnwidth]{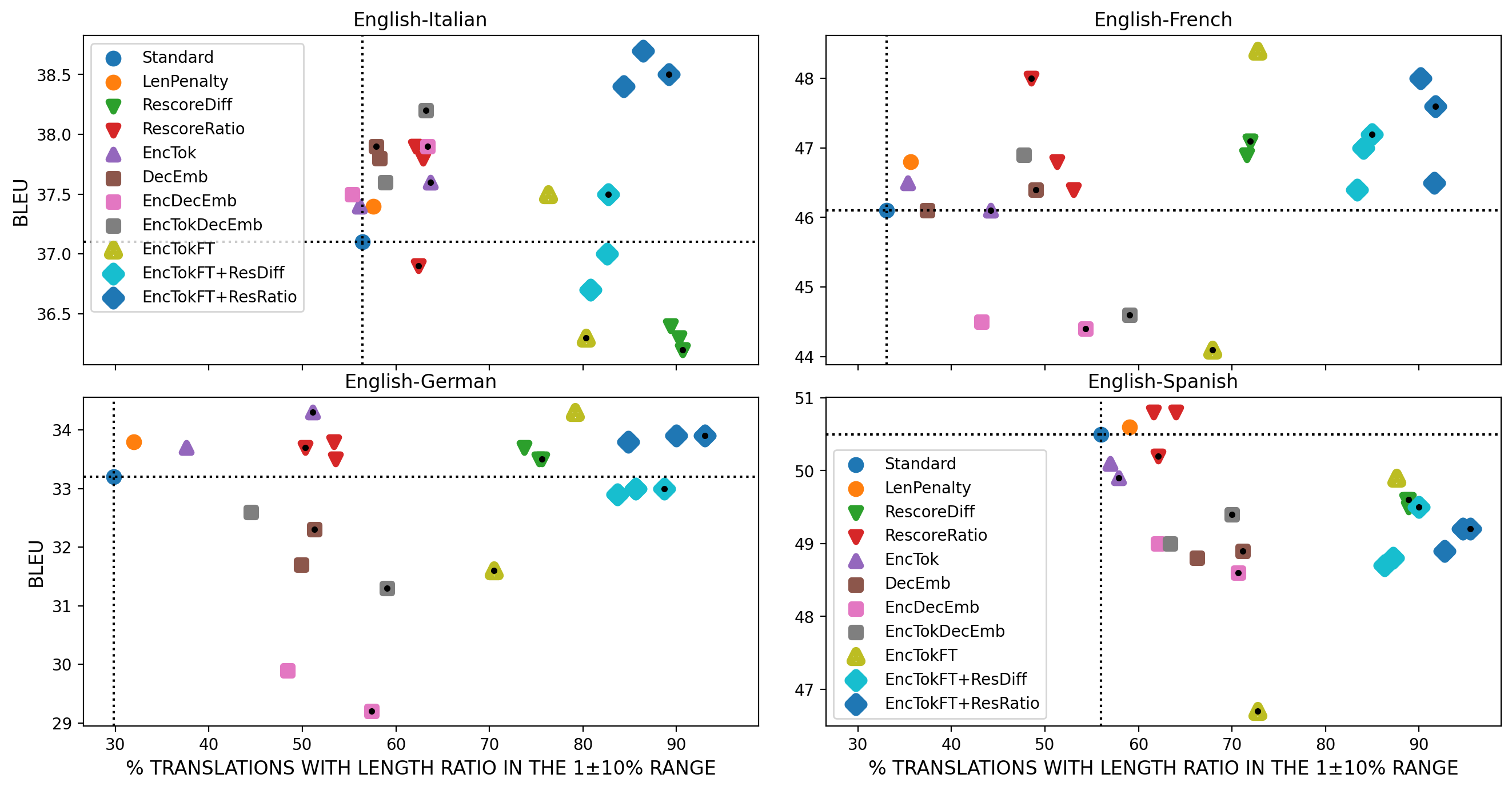}
    \caption{Evaluation of verbosity control approaches in four translation directions. Methods are evaluated with respect to quality (BLEU) and verbosity (\% of translations with acceptable length ratio). Dotted lines indicate performance of the standard MT model: hence, better systems should fall in the top right sector. Model families have distinct symbols: standard models ($\circ$), verbosity embedding ($\square$), verbosity token ($\bigtriangleup$), rescoring models ($\triangledown$),   verbosity token + rescoring ($\Diamond$). Results of {\tt Short} and {\tt Normal} verbose/embedding models are shown, former are marked ($\bullet$). Rescoring results are shown for limited sub-ranges of $\alpha$, by marking ($\bullet$) common $\alpha$ values for all languages.  }
    \label{fig:verb-figs}
\end{figure*}

\section{Experiments}
\subsection{Data and Metrics}
\mf{We evaluate the proposed approaches on four translation directions, English (En) to Italian (It), French (Fr), German (De) and Spanish (Es). We present results on the MuST-C corpus~\cite{MUSTC:2019} of TED talks. 
To simulate realistic production settings, we also leverage proprietary parallel data at the magnitude of $10^7$ sentences for MT model pre-training of each pair. 
Data is first preprocessed with scripts from the Moses~\cite{koehn2007moses} tool.\footnote{Moses: https://github.com/moses-smt/mosesdecoder} We then use SentencePiece~\cite{kudo2018sentencepiece}\footnote{https://github.com/google/sentencepiece} to learn a sub-word model with 32k merge operations, followed by segmentation of the detokenized text version. 

To evaluate MT performance, we re-translated a subset of 620 sentences from the original MUST-C test, by asking our external vendors to produce translations close in length to the source. To measure MT quality, we use the BLEU score computed with  SacreBLEU~\cite{post-2018-call}.\footnote{SacreBLEU: https://github.com/mjpost/sacrebleu} To evaluate verbosity control, we count the $\%$ of outputs meeting the target-source length ratio of $1 \pm 10\%$, which we consider acceptable for AD.}

\noindent
\subsection{Model and Settings} 
\mf{All models are implemented using Transformer~\cite{vaswani2017attention}, with 6 layers of encoder-decoder network. Each layer constitutes sub-layers of self-attention of size 1024 with 16 heads and feed-forward with 4096 hidden dimensions. Adam optimizer~\cite{kingma2014adam} is used with a learning rate of $1\times10^{-7}$, that linearly increases for the first $4000$ steps, followed by a decrease with an inverse square root of the model training steps. A dropout of 0.1 is applied on the attention layer, while 0.3 is used for the rest. All presented systems are trained on the large data set and fine tuned on the in-domain data. At fine-tuning time, employing similar settings has shown better performance, with the only exception of resetting the optimizer state. Training is performed on 8 V100 
GPUs. Fine-tuning is performed for a total of 20 epochs. Models for evaluation are selected based on the validation loss. Except for rescoring models, where we set a beam size of 50 as in~\cite{saboo_integration_2019}, we use a beam size of 5 for all inferences.}

\subsection{Intrinsic Evaluation}
\mf{We first apply the single-stage fine-tuning method with our verbosity models. Experimental results with MT quality and verbosity control scores are in Figure~\ref{fig:verb-figs}. For all language pairs, the standard Transformer model ({\it Standard}) is presented as the reference system to be improved along both dimensions. Our observations are as follows.

First, the length penalty ({\it LenPenalty}) baseline model, representing a naive way of generating shorter translations, surprisingly shows a slight gain both in BLEU and verbosity control with respect to the {\it Standard}.} Second, \mf{N-best rescoring as in ~\cite{saboo_integration_2019} ({\it RescoreDiff}) displays mixed behaviours. Verbosity control on IT and ES reaches 90\%, but BLEU drops, while on DE and FR it improves both dimensions.} N-best rescoring using our proposed synchrony sub-score {\it RescoreRatio}, by contrast, shows a more consistent behaviour across all languages. For instance, it improves BLEU score and source-target length ratio for all language pairs with the {\tt Normal} token.

Third, our verbosity models with {\tt Short} and {\tt Normal}  tokens ({\it EncTok}) improve both metrics across all languages, except for $En-Es$ pair. Moreover, the {\tt Short} token (marked)  tends to have better performance than the {\tt Normal} token (unmarked). This confirms the need of reducing the LR to make translation outputs from \textit{English} to other languages more or less the same as that of the source sentence.

Moreover, in both {\tt Short} and {\tt Normal} settings, the verbosity embedding approaches ({\it DecEmb, EncDecEmb} and {\it EncTokDecEmb} show a less consistent behaviour. In particular, for DE and ES we observe MT quality drops by all such models, while for FR we observe MT quality drops by  {\it EncDecEmb, EncTokDecEmb}. Finally, using the verbosity embedding as output layer bias in the decoder \cite{michel-neubig-2018-extreme} did not provide any improvements in MT quality nor verbosity control with respect to the {\it Standard} model. For the sake of brevity they are not reported in the paper. 

Given its consistent behavior we explore the {\it EncTok} method further. We first investigate the combination of the method and the two-stage fine-tuning procedure, as in Figure~\ref{fig:verb-figs}. By applying the two stage fine-tuning instead of the one-stage fine-tuning method, our verbosity token ({\it EncTokFT}) model with the {\tt Normal} token gains better performance in all languages except Spanish where we see a slight loss in BLEU. In terms of verbosity control, the model translation outputs (unmarked) satisfy the length requirement well over 70$\%$ in all pairs. This verifies the effectiveness of our proposed fine-tuning procedure method in verbosity controlling. That is, the two-stage fine-tuning utilizes better the verbosity token in generating translation more or less the same as that of the source sentence.

Next, we combined the latter model with both N-best rescoring methods ({\it EncTokFT + RescoreDiff} and {\it EncTokFT + RescoreRatio}). We found the
{\tt Normal} setting works better for the combination and thus report 
only these results for the sake of clarity. Both combined methods further improve all previous methods. Moreover, rescoring using our poposed synchrony sub-score {\it RescoreRatio} is often better regarding both metrics of BLEU and source target length ratio. For instance {\it EncTokFT + RescoreRatio} pushes the percentage of acceptable length well over 90$\%$ for three pairs and to $89.9\%$ for IT. In addition, {\it EncTokFT + RescoreRatio} improves BLEU score for all languages, but for Spanish (-2.5\% relative BLEU):  +4\% for IT (+1.4 BLEU), +3\% for FR (+1.5 BLEU) and +2\% for DE (+0.70 BLEU).

\subsection{Extrinsic Evaluation}
We run an extrinsic subjective evaluation on a subset of 120 sentences to directly  measure the impact that verbosity control of MT has on speech dubbing. In particular, we limit the comparison to Italian and German and to translations generated with {\it Standard} and our best model {\it EncTokFT + RescoreRatio}. After removing identical translations by the two systems, we end up with 100 and 110 sentences, respectively. We generate dubbed videos in Italian and German from them by using the architecture described in Section 2. As a reference, we also generate dubs from the reference translations. We split the test into batches that we assigned to a total of 40 subjects (proportions vary by language). Subjects were asked to watch the reference video and then rate their user experience with the two dubbing variants which were presented in random order and anonymously on a scale from 0 to 10 (higher is better).

We collected a total of 2,000 and 2,200 judgments for each variant and used them for a head-to-head comparison. By looking at the percentage of wins in Italian: {\it EncTokFT + RescoreRatio} got 38.7\% wins against {\it  Standard} got 32.45\% ($p<0.01$) (the rest were ties). For German, {\it EncTokFT + RescoreRatio} got 40.0\% wins against {\it  Standard} got 
33.64\% ($p<0.02$) (the rest were ties).

\section{Conclusions}
\hc{We have presented and systematically compared verbosity control methods of MT in order to generate translations of length that is appropriate for automatic dubbing. Our analysis includes  methods of integrating verbosity tokens and embeddings, fine-tuning strategies with verbosity information and finally rescoring functions to select outputs with \sml{the desired} quality and verbosity. Compared to a standard Transformer MT model trained without verbosity information, our resulting best model  not only produces translations much closer in length to the input, but often also better in translations. We also conducted a subjective evaluation on automatically dubbed videos using the translations generated by MT with and without verbosity control. The results confirm an increase in human preference for videos dubbed with the latter version.}



\ninept
\bibliographystyle{ieeetr}
\bibliography{paper}
\end{document}